\documentclass[11pt]{article}

\usepackage[final]{acl}

\usepackage{times}
\usepackage{latexsym}

\usepackage[T1]{fontenc}

\usepackage[utf8]{inputenc}

\usepackage{microtype}

\usepackage{inconsolata}

\usepackage{graphicx}

\usepackage{algorithm}
\usepackage{algorithmic}
\usepackage{booktabs}
\usepackage{multirow}
\usepackage{siunitx}
\usepackage{makecell}
\usepackage{amsmath}
\usepackage{amssymb}
\usepackage{pifont}
\usepackage{CJKutf8}

\title{Decoding the Multimodal Mind: Generalizable Brain-to-Text Translation via Multimodal Alignment and Adaptive Routing}

\author{
 \textbf{Chunyu Ye\textsuperscript{1,2}},
 \textbf{Yunhao Zhang\textsuperscript{1,2}},
 \textbf{Jingyuan Sun\textsuperscript{3}},
\\
 \textbf{Chong Li\textsuperscript{1,2}},
 \textbf{Yang Zhao\textsuperscript{1,2}},
 \textbf{Shaonan Wang\textsuperscript{4}\thanks{Corresponding author.}},
\\
\\
 \textsuperscript{1}State Key Laboratory of Multimodal Artificial Intelligence System,\\Institute of Automation, Chinese Academy of Sciences\\
 \textsuperscript{2}School of Artificial Intelligence, University of Chinese Academy of Sciences\\
 \textsuperscript{3}Department of Computer Science, The University of Manchester\\
 \textsuperscript{4}Department of Language Science and Technology, Hong Kong Polytechnic University
\\
\normalfont \texttt{yechunyu2001@outlook.com},
\normalfont \texttt{shaonan.wang@polyu.edu.hk}
}

\begin{document}
\maketitle
\begin{abstract}
Decoding language from the human brain remains a grand challenge for Brain-Computer Interfaces (BCIs). Current approaches typically rely on unimodal brain representations, neglecting the brain's inherently multimodal processing. Inspired by the brain's associative mechanisms, where viewing an image can evoke related sounds and linguistic representations, we propose a unified framework that leverages Multimodal Large Language Models (MLLMs) to align brain signals with a shared semantic space encompassing text, images, and audio. A router module dynamically selects and fuses modality-specific brain features according to the characteristics of each stimulus. Experiments on various fMRI datasets with textual, visual, and auditory stimuli demonstrate state-of-the-art performance, achieving an 8.48\% average improvement on the most commonly used benchmark. We further extend our framework to EEG and MEG data, demonstrating flexibility and robustness across varying temporal and spatial resolutions. To our knowledge, this is the first unified BCI architecture capable of robustly decoding multimodal brain activity across diverse brain signals and stimulus types, offering a flexible solution for real-world applications.
\end{abstract}

\section{Introduction}

\begin{figure}[t]
\centering
\includegraphics[width=0.9\columnwidth]{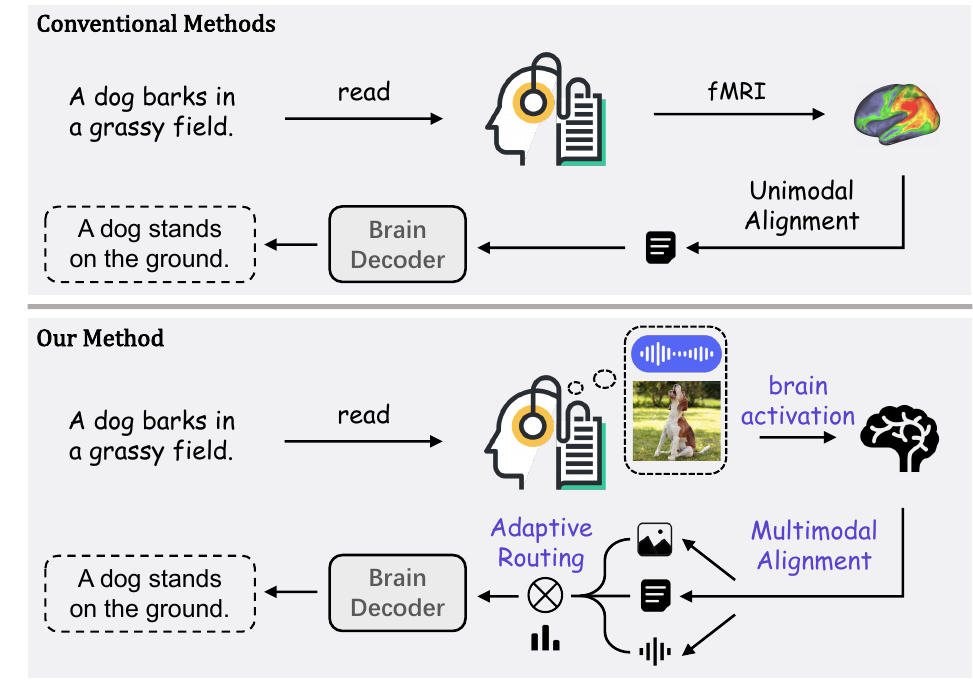} 
\caption{Top: Conventional BCI frameworks decode language from unimodal brain representations. Bottom: Our proposed framework (supports diverse brain signals and stimulus types) leverages multimodal brain representations for language decoding via multimodal alignment and adaptive routing strategy.}
\label{fig1}
\end{figure}

Brain-Computer Interfaces (BCIs) enable users to interact with the external world using only their thoughts. One of the most significant applications of BCIs is language decoding, which aims to translate brain signals into text. This capability holds tremendous promise for restoring communication in individuals with severe language impairments, thereby reconnecting them with their environment. However, reliable language decoding remains exceptionally challenging because the brain representation of thought is inherently complex and multimodal \cite{meyer2007hemodynamic, barsalou2008grounded, anderson2017visually}. For example, reading the sentence ``a dog barks in a grassy field'' engages not only linguistic processing but also evokes vivid visual and auditory imagery. Consequently, recorded brain signals comprise a rich mixture of linguistic, visual, and auditory representations.

Most existing decoding models, however, exploit only a single modality of these brain representations \cite{mai2023unibrain,ferrante2023brain,xia2024umbrae,qiu2025mindllm}, learning a direct mapping from brain signals to a unimodal representation. This strategy cannot generalize to inputs from other modalities and, more importantly, overlooks the brain’s inherently multimodal processing: text can evoke visual imagery, and images can activate language-related brain regions \cite{zhao2025memory,spence2011crossmodal,bolam2022neurocomputational}. We hypothesize that such cross-modal activations provide complementary information that is crucial for effective brain decoding.

To incorporate this associative mechanism into brain decoding, we propose a unified multimodal framework (Figure \ref{fig1}). First, we perform multimodal alignment by learning a mapping from brain signals to a shared semantic space that encompasses text, image, and audio representations. We then introduce an adaptive routing mechanism that dynamically selects and fuses modality-specific brain features for each sample, emulating the brain’s cross-modal associative activation and context-dependent integration. Given a brain sample, the router analyzes the underlying neural representations and determines the contribution of each modality to decode the user’s thoughts. Together, these components enable a unified BCI decoding framework that leverages multimodal activations elicited by unimodal stimuli.

We evaluate the proposed framework on three widely used functional magnetic resonance imaging (fMRI) datasets with text, image, and audio stimuli, achieving state-of-the-art performance on all benchmarks. Notably, we observe an 8.48\% average improvement on the most commonly used dataset. In contrast to previous approaches that are limited to a single dataset or neuroimaging modality, our model demonstrates strong generalizability. We extend the framework to electroencephalography (EEG) and magnetoencephalography (MEG) data, which offer high temporal resolution compared to the high spatial resolution of fMRI, and consistently observe superior performance. Our contributions are summarized as follows:

\begin{itemize}
\item We propose a unified BCI decoding framework that maps brain signals to a shared semantic space across text, image, and audio modalities, enabling effective and scalable brain decoding.
\item We introduce a modality routing mechanism inspired by associative learning that dynamically adjusts the contribution of textual, visual, and auditory representations for each sample.
\item We evaluate the proposed framework across multiple neuroimaging modalities (fMRI, EEG, and MEG) and multiple stimuli (vision, language, and audio), achieving state-of-the-art performance and demonstrating robustness and generalizability.
\end{itemize}

Together, our work presents the first unified and scalable solution for decoding the brain's multimodal representations, advancing toward more robust, adaptive, and general-purpose brain-computer interfaces.

\section{Related work}
This study focuses on decoding text from brain signals (e.g., fMRI) by leveraging the multimodal representation capabilities of a Multimodal Large Language Model (MLLM). The following sections will respectively introduce related work in brain-conditioned text decoding and MLLMs.

\textbf{Brain-Conditioned Text Decoding} is a key component of BCIs, aiming to reconstruct text from brain signals. Early research primarily employed fMRI due to its high spatial resolution and non-invasive nature. These pioneering efforts often utilized classification tasks to extract semantic information from fMRI data \cite{mitchell2008predicting,palatucci2009zero}, such as predicting target words from neural signals using contextual prompts \cite{zou2022cross}. With the advent of large language models (LLMs), fMRI decoding has evolved from a classification paradigm to a continuous generation paradigm. Approaches in this new paradigm include methods that reconstruct text by directly comparing brain-predicted and candidate text embeddings \cite{tang2023semantic,zhao2024mapguide}. Furthermore, BrainLLM \cite{ye2025generative} proposed a method for continuous text decoding by employing fMRI embeddings as a prefix to guide the text generation process.

Moreover, breakthroughs in MLLMs have further expanded the capabilities of fMRI decoding by leveraging additional image information. BrainChat \cite{huang2025brainchat} utilized contrastive learning to align fMRI signals with multimodal (text and image) representations, employing a cross-attention model to effectively decode neural signals. Similarly, UMBRAE \cite{xia2024umbrae} aligned fMRI embeddings with image embeddings to leverage the native image-understanding capabilities of MLLMs, further demonstrating the efficacy of integrating multimodal models into fMRI decoding. However, prior work has often focused on single-stimulus fMRI decoding and faced challenges in effectively utilizing auxiliary information from multiple modalities to aid fMRI decoding.

Meanwhile, decoding efforts based on MEG and EEG data have also been actively pursued. In recent years, researchers have begun harnessing the powerful representational and generalization capabilities of LLMs to enhance MEG/EEG decoding, enabling more accurate mapping of neural signals to linguistic outputs \cite{yang2024mad,liu2024eeg2text,duan2023dewave}. However, most of these studies focus exclusively on a single type of brain signal and rely solely on unimodal representations, which limits both the generalizability of BCIs and the accuracy of decoding. Here, we propose a unified model that processes brain signals from fMRI, MEG, and EEG by aligning brain embeddings with multimodal embeddings to generate coherent text.

\textbf{Multimodal Large Language Models} are designed to process and understand multimodal information through various pretraining tasks such as contrastive learning, masked language modeling, and image-text matching. Most work in this area has focused on image and text. Flamingo \cite{alayrac2022flamingo} was an early attempt that infused frozen visual features into LLMs via cross-attention, enabling the model to achieve excellent performance across various multimodal benchmarks. With the rise of instruction-following LLMs, models like Qwen2-VL \cite{wang2024qwen2} connect a pre-trained CLIP-ViT \cite{radford2021learningtransferablevisualmodels} and a Qwen LLM \cite{yang2024qwen2technicalreport} through a linear layer, enabling it to follow visual instructions and perform general-purpose visual and language understanding.

Recent efforts have extended beyond vision to include additional modalities such as audio and video. Qwen2.5-Omni \cite{xu2025qwen2}, for example, is a streaming end-to-end multimodal model that integrates text, image, audio, and video using block-wise encoding and time-aligned position embeddings for synchronized cross-modal understanding and generation. Other efforts, such as OneLLM \cite{han2024onellm}, have begun exploring unconventional modalities like point clouds and fMRI. However, these models typically require significant computational resources and large-scale training data. In this work, we leverage Qwen2.5-Omni's multimodal representation capabilities and introduce an efficient alignment strategy that enables learning from minimal supervision with only a few training examples.

\begin{figure*}[t]
\centering
\includegraphics[width=0.8\textwidth]{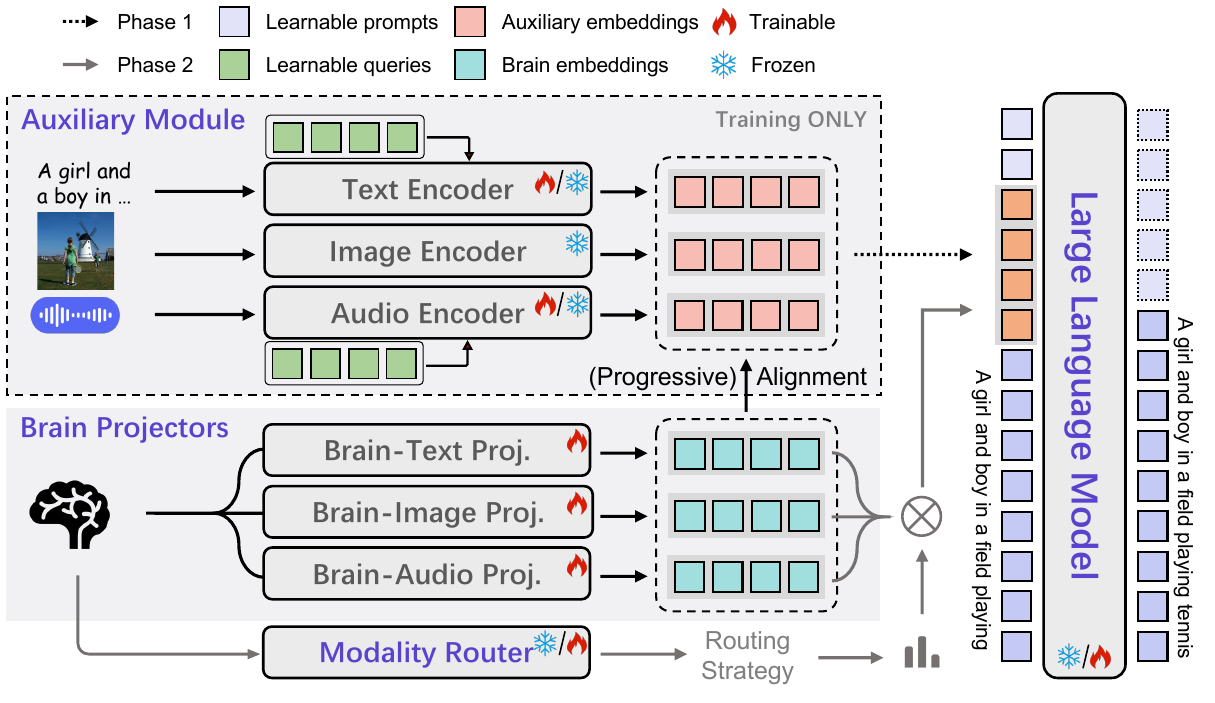}
\caption{Architecture of the proposed BCI framework, consisting of four key components: an Auxiliary Module (providing auxiliary embeddings for alignment during training), Brain Projectors (for brain feature extraction), a Modality Router (for fusing multimodal brain features), and an LLM (for text generation). Training involves two phases: (1) Multimodal Instruction Tuning: The Auxiliary Module is trained to produce modality-specific embeddings for multimodal captioning, with Brain Projectors concurrently trained to align them; (2) Projectors Fusion: the Modality Router learns to dynamically combine projector outputs for LLM-driven text generation. During inference, brain signals (fMRI, EEG, or MEG) are directly projected into the multimodal brain embeddings and fused to generate coherent text without any auxiliary information.}
\label{framework}
\end{figure*}

\section{Method}
We propose a unified brain decoder, as illustrated in Figure \ref{framework}, which achieves high-performance text decoding through multimodal alignment and adaptive routing strategy. In this section, we present the overall model architecture (Sec. \ref{sec:core_components}), and describe the two training phases: multimodal instruction tuning (Sec. \ref{sec:mit}) and projectors fusion (Sec. \ref{sec:projector_fusion}). During training, the decoder aligns brain signals with associated auxiliary multimodal information, while during inference, it directly generates text from brain signals.

\subsection{Core components} \label{sec:core_components}

\textbf{Auxiliary Module} generates modality-specific embeddings from image, text, or audio inputs, which are then utilized by the LLM for multimodal captioning. Concurrently, these embeddings also serve as auxiliary embeddings for training the Brain Projectors. The text encoder employs a transformer architecture with a cross-attention module, where learnable queries attend to pre-trained text embeddings (keys/values) to yield a fixed-length vector. The image encoder, based on CLIP-ViT \cite{radford2021learningtransferablevisualmodels}, extracts effective semantic information. The audio encoder combines Whisper-large-v3 \cite{radford2023robust} with a cross-attention module to transform audio signals via learnable queries into fixed-length vectors. All three modalities produce embeddings of identical dimensionality within a unified representation space, facilitating Brain Projectors alignment. This module is exclusively used during training to enhance alignment and is not employed during inference.

\textbf{Brain Projectors} consist of $M$ projectors, denoted $\{P_1, P_2, \dots, P_M\}$, each based on the CLIP-ViT architecture. The number of Brain Projectors matches the $M$ auxiliary modalities processed by the Auxiliary Module, ensuring a one-to-one mapping. Each projector $P_i$ encodes brain signals into distinct embeddings, which are then aligned with the corresponding auxiliary embedding.

\begin{figure}[t]
\centering
\includegraphics[width=0.95\columnwidth]{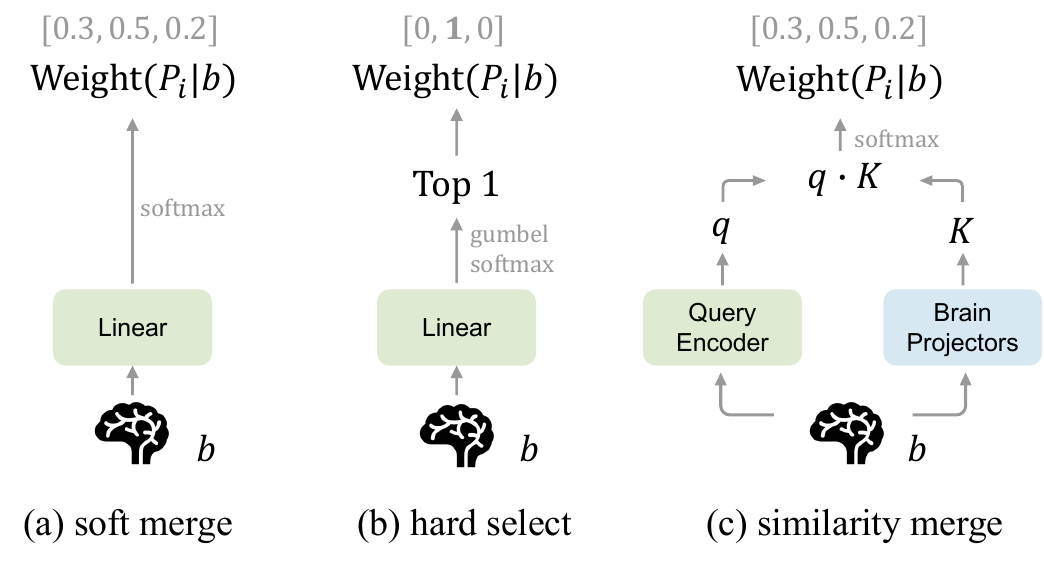}
\caption{Comparison of three routing strategies for calculating Brain Projector importance weights. (a) Soft merge: a linear layer with Softmax calculates weights for all projectors. (b) Hard select: a linear layer with Gumbel-Softmax selects the top-scoring projector. (c) Similarity merge: dot‐product similarities between the query and brain embeddings are Softmax‐normalized to serve as weights.}
\label{router}
\end{figure}

\textbf{Modality Router} Unlike previous approaches that treat the brain as unimodal, we introduce the Modality Router, inspired by the brain's associative processes, which calculates the contribution of different modalities (e.g., visual, textual, auditory) for each brain sample, leveraging the multimodal nature of brain representation. Given a raw brain sample $b$, the Modality Router $\mathcal{R}$ assigns a weight $w_i$ to each Brain Projector $P_i$, facilitating the dynamic integration of multimodal information. We propose three distinct routing strategies for the Modality Router, as illustrated in Figure \ref{router}.

The soft merge strategy, shown in Figure \ref{router}(a), constructs the router as a multi-layer perceptron (MLP). The brain sample $b$ is fed through the MLP and a subsequent softmax function to yield a probability distribution over the modalities. 

In contrast, the hard select strategy, depicted in Figure \ref{router}(b), selects the single most relevant modality for each brain sample. Like the soft merge strategy, it employs an MLP; however, to enable differentiable hard selection during training, we employ the Gumbel-Softmax function \cite{jang2016categorical}. Specifically, given the logits $\mathbf{l} = \mathrm{MLP}(b)$, the hard selection is approximated by:
\begin{equation}
y_i = \mathrm{softmax}\left(\frac{l_i + g_i}{\tau}\right),
\end{equation}
where $l_i$ denotes the $i$-th logit, $g_i\sim\mathrm{Gumbel}(0,1)$, and $\tau > 0$ is a temperature hyperparameter that controls the sharpness of the distribution. A Top-1 selection is then applied to transform the continuous vector $\boldsymbol{y}$ into a one-hot weight vector $\boldsymbol{w}$:
\begin{equation}
w_i =
\begin{cases}
1, & i = \arg\max_j y_j,\\
0, & \text{otherwise.}
\end{cases}
\end{equation}

During backpropagation, gradients are computed with respect to the continuous values $\boldsymbol{y}$, enabling end-to-end optimization via the \emph{straight-through estimator}. This approach allows the model to make discrete modality selections while preserving differentiability.

Finally, the similarity merge strategy (Figure \ref{router}(c)) utilizes an attention mechanism to assign modality weights. A query encoder processes the raw brain signal $b$ to produce a query vector $q$. Concurrently, the Brain Projectors provide their respective multimodal brain embeddings $P_i(b)$, forming the key set $K = \{k_1, k_2, \dots, k_M\}$. The weight $w_i$ for each modality is obtained from the similarity between the query vector $q$ and each modality's key vector $k_i$, computed using the dot product followed by softmax normalization $w_i = \mathrm{softmax}(q \cdot k_i)$. This procedure yields context-sensitive modality weights that capture the learned relevance of each representation to the brain-derived query.

All three routing strategies ultimately compute weights $w_i$ for each projector, which are then used to fuse the multimodal brain embeddings as
\begin{equation}
H = \sum w_i P_i(b).
\label{eq:fusion}
\end{equation}

\textbf{LLM} We adopt the language decoder of Qwen2.5-Omni \cite{xu2025qwen2} as the LLM. Its input is a sequence of learnable soft-prompt tokens followed by either a brain or modality embedding. The soft-prompt is initialized using pre-trained embeddings derived from instruction text. This approach is motivated by studies suggesting that soft prompts aligned with the pre-trained embedding distribution of the LLM tend to facilitate more effective convergence and enhanced performance \cite{lester2021power}.

\subsection{Multimodal Instruction Tuning} \label{sec:mit}

We first employ prompt tuning to optimize the modality encoders within the Auxiliary Module. This process extracts fixed-length, modality-specific embeddings that serve as auxiliary embeddings. Subsequently, the Brain Projectors align brain embeddings to these auxiliary embeddings, ensuring effective cross‑modal correspondence.

The Auxiliary Module is optimized to produce modality-specific embeddings that, when fed into the LLM, enable multimodal captioning. Specifically, given auxiliary information paired with brain signals, the LLM is required to generate the corresponding textual description. Concretely, for the image modality it performs standard image captioning; for text, input re‑paraphrasing; and for audio, speech transcription. This objective encourages each encoder to learn semantically rich representations aligned with the LLM's embedding space. The captioning loss is defined as
\begin{equation}
\mathcal{L}_{\mathrm{cap}}
\;=\; -\sum_{k=1}^{T} \log P_{\theta}\bigl(t_k \mid t_{<k},\,[S; {z_m}]\bigr),
\end{equation}
where $T$ is the target length, $t_k$ the $k$-th token with prefix $t_{<k}$, $[S;z_m]$ the concatenation of soft-prompt $S$ and modality embedding $z_m$, and $P_\theta$ the LLM's distribution.

Once the Auxiliary Module produces embeddings sufficient for multimodal captioning, we transfer this knowledge to the Brain Projectors through alignment. We employ Mean Squared Error (MSE) for this purpose. The alignment loss encourages each brain embedding $z_b$ to match its corresponding modality-specific auxiliary embedding $z_{m}$:
\begin{equation}
\mathcal{L}_{\mathrm{align}}
= \mathbb{E}_{(z_b,z_m)\sim\mathcal{D}}\!\bigl[\lVert z_b - z_m\rVert_{2}^{2}\bigr].
\end{equation}

Although the image encoder is pre‑aligned to the LLM space, the text and audio encoders are randomly initialized and require training to produce effective embeddings. Therefore, we employ a dynamic weighting scheme for progressive alignment:
\begin{equation}
\alpha(t) \;=\; \frac{1}{1 + e^{-\lambda\,(t - t_{0})}},
\end{equation}
where $\lambda$ controls the transition sharpness, $t_{0}$ denotes the midpoint step, and $t$ is the current training iteration.

Ultimately, we jointly minimize the captioning and alignment losses according to a progressive schedule:
\begin{equation}
\mathcal{L}(t) \;=\; \mathcal{L}_{\mathrm{cap}}
\;+\;\alpha(t)\,\mathcal{L}_{\mathrm{align}}.
\end{equation}

At the start of training, $\alpha(t)\approx 0$ biases optimization towards multimodal captioning; as $t$ increases, $\alpha(t)\to 1$ gradually shifts focus to encoder-projector alignment. Upon completion, brain and auxiliary embeddings are well aligned within the LLM's representation space.

\subsection{Projectors Fusion} \label{sec:projector_fusion}

After Multimodal Instruction Tuning aligns brain embeddings with specific modalities, this phase seeks to effectively fuse these embeddings to better leverage the multimodal nature of brain signals. We route brain signals through a Modality Router to compute contribution scores $w_i$ for each modality, then weight the Brain Projector outputs $P_i(b)$ to obtain a unified embedding (See Eq. \eqref{eq:fusion}). This multimodal embedding, $H$, is subsequently passed into the LLM for language modeling.

To prevent overfitting to a single projector and promote balanced training, we introduce a sample-level load balancing strategy for both merge-based routers (soft and similarity merge) and select-based routers (hard select):

\begin{equation}
\mathcal{L}_{\mathrm{balance}}
=
\begin{cases}
\displaystyle
-\frac{1}{N}
\sum_{i=1}^{N}\sum_{k=1}^{M}\log w_{i,k},
&\text{(merge)}\\[1.5ex]
\displaystyle
M \sum_{k=1}^{M} f_{k}\,P_{k},
&\text{(select)}
\end{cases}
\end{equation}

where $N$ is the batch size, $M$ the number of projectors, and $w_{i,k}$ the weight assigned to projector $k$. For merge‑based routers, the logarithmic term prevents any projector's weight from vanishing. For select‑based routers, we define:
\begin{equation}
f_k = \frac{1}{N}\sum_{i=1}^{N}\mathbb{I}\bigl[\mathrm{proj}(b_i)=k\bigr],
\quad
P_k = \frac{1}{N}\sum_{i=1}^{N} w_{i,k},
\end{equation}
where $f_k$ is the fraction of samples routed to projector $k$ and $P_k$ its average gating probability. Thus, the merge‑based term maximizes the product of total weights across projectors, while the select‑based term aligns empirical routing frequencies $f_k$ with gating probabilities $P_k$.

Through the load balancing loss, the brain router can more comprehensively utilize the multimodal brain representations. 
To prevent the distribution differences from expanding further in this training step, we still introduce the alignment loss to regularize the distribution of the brain embeddings. The final loss is
\begin{equation}
\mathcal{L} = \mathcal{L}_{\mathrm{cap}}
+ \lambda_1\,\mathcal{L}_{\mathrm{align}}
+ \lambda_2\,\mathcal{L}_{\mathrm{balance}},
\end{equation}
where $\lambda_1$ and $\lambda_2$ are hyperparameters that weight the alignment loss $\mathcal{L}_{\mathrm{align}}$ and the load balancing loss $\mathcal{L}_{\mathrm{balance}}$, respectively.

\section{Experimental Setup}
\subsection{Datasets} For the fMRI data, we utilized three different modality-specific datasets induced by visual (NSD), textual (Pereira), and auditory (Huth) stimuli. The NSD dataset \cite{allen2022massive} provides paired fMRI and image data based on COCO stimuli, using standard train/test splits \cite{huang2025brainchat, xia2024umbrae}. The Pereira dataset \cite{pereira2018toward} comprises fMRI recordings collected while participants read encyclopedic sentences; each continuous scan captures brain activity elicited by multiple words simultaneously. The Huth dataset \cite{lebel2023natural} contains fMRI data from 8 participants listening to 27 natural narrative stories, with official preprocessing applied.

The SMN4Lang MEG dataset \cite{wang2022synchronized} comprises recordings from 12 participants listening to 60 stories. The ZuCo2 EEG dataset \cite{zou2022cross} includes EEG responses from 18 participants to 739 Wikipedia sentences. More details about the datasets are provided in Appendix \ref{app:dataset_details}.

\subsection{Baselines}
For all datasets, we establish a lower bound by pairing mismatched brain signals with the target text. We also implement a simple baseline that encodes brain signals with a ViT encoder followed by a linear layer before feeding them into an LLM.

For the NSD dataset, we compare models trained solely on single-subject data to ensure a fair evaluation (See Appendix \ref{app:other_sub} for the results of other subjects). The following baselines are considered: SDRecon \cite{takagi2023high}, UniBrain \cite{mai2023unibrain}, and BrainCap \cite{ferrante2023brain}, which primarily utilize linear regression to map fMRI to LLM input space. OneLLM \cite{han2024onellm} stands out as an MLLM that aligns fMRI, alongside seven other modalities, with language. BrainChat \cite{huang2025brainchat} employs contrastive cross‑attention to align fMRI with text. UMBRAE \cite{xia2024umbrae} maps fMRI embeddings into the same space as image embeddings to leverage MLLMs' native image‑understanding capabilities. MindLLM \cite{qiu2025mindllm} employs a neuroscience-informed attention mechanism for fMRI decoding. Notably, all baseline methods consider only unimodal brain representations, whereas our model effectively exploits the brain’s multimodal capabilities (See Appendix \ref{app:method_comp}).

For other datasets, most of the methods mentioned above were not applied due to their heavy reliance on visual stimuli. We therefore use BrainLLM \cite{ye2025generative} as a principal baseline due to its intuitive and broadly generalizable framework. For the Huth dataset, we include an additional competitive baseline proposed by \citet{tang2023semantic}.

\begin{table*}[ht]
\centering
\begingroup
\setlength{\tabcolsep}{1.0mm}
\resizebox{\textwidth}{!}{
\begin{tabular}{@{}lccccccccc@{}}
\toprule
Method    & BLEU-1 & BLEU-2 & BLEU-3 & BLEU-4 & METEOR & ROUGE-L  & CIDEr   & CLIP-S  & RefCLIP-S\\
\midrule
mismatched   &34.21 &15.62 &6.45 &2.89 & 9.18 & 23.68 & 12.05 & 59.72 & 65.15\\
linear & 46.21 & 25.45 & 13.88 & 8.62 & 12.72 & 32.45 & 21.12 & 58.45 & 63.82\\
SDRecon   &36.21 &17.11 &7.22 &3.43 &10.03 &25.13 &13.83 &61.07 &66.36\\
OneLLM    & 47.04 & 26.97 & 15.49 & 9.51  & 13.55 & 35.05 & 22.99     & 54.80   & 61.28\\
UniBrain  & -     & -     & -     & -     & 16.90 & 22.20 & -         & -       & -    \\
BrainCap  & 55.96 & 36.21 & 22.70 & 14.51 & 16.68 & 40.69 & 41.30     & \underline{64.31}   & 69.90\\
BrainChat & 52.30 & 29.20 & 17.10 & 10.70 & 14.30 & 45.70 & 26.10     & -       & -    \\
UMBRAE    & 57.63 & 38.02 & 25.00 & 16.76 & 18.41 & 42.15 & 51.93     & \textbf{66.44}   & \textbf{72.12} \\
MindLLM  & 58.05 & 37.95 & 24.40 & 16.14 & 16.62 & 42.03 & 43.04     & -   & - \\
\midrule
Ours (soft merge) & \underline{61.46} & \underline{43.37} & \underline{30.07} & \underline{21.04} & \underline{20.60} & \underline{46.15} & \underline{63.68} & 62.72 & 69.72 \\
Ours (hard select) & 60.69 & 42.59 & 29.90 & 21.33 & 19.67 & 45.66 & 59.55 & 60.31 & 67.75 \\
Ours (similarity merge) &\textbf{61.53}	&\textbf{43.45}	&\textbf{30.21}	&\textbf{21.92}	&\textbf{20.71}	&\textbf{46.45}	&\textbf{63.88}	&63.13	&\underline{70.12} \\
\bottomrule
\end{tabular}}
\caption{Results of brain captioning on NSD subject 1. \textit{Soft merge}, \textit{hard select}, and \textit{similarity merge} refer to the three routing strategies illustrated in Figure \ref{router}.}
\label{tab:NSD_main}
\endgroup
\end{table*}

\begin{table*}[ht]
\centering
\begingroup
\setlength{\tabcolsep}{1.0mm}
\resizebox{\textwidth}{!}{
  \begin{tabular}{%
    l  
    l  
    *{4}{c}  
    c@{\quad}  
    *{4}{c}  
  }
    \toprule
    \multirow{2}{*}{Dataset} & \multirow{2}{*}{Method}
      & \multicolumn{4}{c}{with text prompt} 
      & & \multicolumn{4}{c}{without text prompt} \\
    \cmidrule(lr){3-6} \cmidrule(lr){8-11}
    & 
      & BLEU-1$\uparrow$ & ROUGE-1$\uparrow$ & ROUGE-L$\uparrow$ & WER$\downarrow$
      & & BLEU-1$\uparrow$ & ROUGE-1$\uparrow$ & ROUGE-L$\uparrow$ & WER$\downarrow$ \\
    \midrule
    \multirow{4}{*}{Pereira}
      & mismatched                 & 31.29 & 24.15 & 24.01 & 83.38 
                               & &  7.87 &  5.53 &  5.40 & 97.26 \\
      & linear                 & 32.48 & 25.21 & 25.13 & 82.15 
                               & &  9.21 &  7.12 &  7.04 & 96.92 \\
      & BrainLLM               & \underline{33.51} & \underline{26.99} & \underline{26.38} & \underline{81.03}
                               & & \underline{10.25} &  \underline{7.88} &  \underline{7.49} & \underline{96.10} \\
      & Ours                   & \textbf{36.32} & \textbf{28.31} & \textbf{27.82} & \textbf{79.32}
                               & & \textbf{14.98} & \textbf{8.31} & \textbf{8.15} & \textbf{95.69} \\
    \midrule
    \multirow{5}{*}{Huth}
      & mismatched                 & 14.89 & 13.37 & 13.24 & 93.35
                               & &  9.60 &  8.17 &  7.79 & 97.03 \\
      & linear                 & 15.32 & 14.12 & 14.21 & 93.01
                               & &  9.62 &  8.17 & 7.81 & 97.11 \\
      & \citet{tang2023semantic} & 14.95 & 13.39 & 13.26 & 93.34
                               & &  9.67 &  8.18 &  7.88 & 97.00 \\
      & BrainLLM               & \underline{16.91} & \underline{15.81} & \underline{15.03} & \underline{92.16}
                               & & \underline{13.56} & \underline{11.60} & \underline{10.99} & \underline{95.41} \\
      & Ours                   & \textbf{17.35} & \textbf{15.89} & \textbf{15.28} & \textbf{91.35}
                               & & \textbf{14.72} & \textbf{11.79} & \textbf{11.31} & \textbf{95.03} \\
    \bottomrule
  \end{tabular}}
\caption{Results of brain captioning on Pereira and Huth dataset. \textit{With text prompt} indicates that in addition to brain embeddings, we also provide textual context as input to the LLM to generate the corresponding text.}
\label{tab:pere_huth}
\endgroup
\end{table*}

\subsection{Metrics}
We evaluate the generated captions using standard text-based metrics, including BLEU-$k$ \cite{papineni2002bleu}, METEOR \cite{banerjee2005meteor}, ROUGE \cite{lin2004rouge}, CIDEr \cite{vedantam2015cider}, and word error rate (WER). For datasets with reference images, we additionally assess caption–image  relevance using CLIP-S and RefCLIP-S \cite{hessel2021clipscore}. Together, these metrics provide a comprehensive assessment of both linguistic fidelity and visual alignment.

\section{Results}
We evaluated our model on three fMRI datasets and extended its application to MEG and EEG data. Section \ref{sec:quantitative_results} presents the quantitative results. Section \ref{sec:analysis} examines how our model integrates different modalities and provides an interpretable analysis of the effectiveness of multimodal fusion. Finally, Section \ref{sec:ablation} reports ablation studies on the utilized modalities. Examples of the model's decoding results are provided in Appendix \ref{app:examples}.

\subsection{Quantitative Results} \label{sec:quantitative_results}

\textbf{fMRI} As shown in Table \ref{tab:NSD_main} on the NSD dataset, our model outperforms those trained exclusively on unimodal brain representations, achieving an average improvement of 8.48\%. This result validates the effectiveness of our framework. UMBRAE attains the best performance on the CLIP-based metric, likely because it directly optimizes brain-image alignment, resulting in embeddings that capture more image-specific information. However, its performance on text-based metrics is comparatively lower. In contrast, our approach yields a more balanced and generally superior average performance.

On both the Pereira and Huth datasets, as shown in Table \ref{tab:pere_huth}, our model achieves the best performance across most metrics compared to all baselines. We further analyzed the scenario in which the text prompt is not provided as contextual input for LLM generation. In this more challenging setting, all metrics decline; however, the performance trends observed with text prompts persist, and our approach still achieves the best results on the majority of metrics.

\textbf{MEG\&EEG} We further applied our model to MEG and EEG datasets, achieving the highest performance across all evaluation metrics (see Table \ref{tab:meg_eeg}) and demonstrating superior generalizability and scalability compared to other methods.

\begin{table}[ht]
\centering
\begingroup
\setlength{\tabcolsep}{0.3mm}
\resizebox{\columnwidth}{!}{
\begin{tabular}{llcccc}
\toprule
Dataset & Method & BLEU-1$\uparrow$ & ROUGE-1$\uparrow$ & ROUGE-L$\uparrow$ & WER$\downarrow$ \\
\midrule
\multirow{4}{*}{MEG}
  & mismatched       & 15.24 & 11.46 & 11.39 & 93.01 \\
  & linear       & 18.98 & 14.26 & 14.38 & 89.94 \\
  & BrainLLM    & \underline{19.59} & \underline{14.78} & \underline{14.32} & \underline{89.10} \\
  & Ours   & \textbf{20.96} & \textbf{16.97} & \textbf{16.58} & \textbf{86.20} \\
\midrule
\multirow{4}{*}{EEG}
  & mismatched       & 16.92 & 12.01 & 11.97 & 92.91 \\
  & linear       & 18.03 & 13.15 & 13.28 & 91.32 \\
  & BrainLLM    & \underline{18.29} & \underline{13.30} & \underline{13.22} & \underline{90.28} \\
  & Ours   & \textbf{20.03} & \textbf{15.93} & \textbf{15.65} & \textbf{89.19} \\
\bottomrule
\end{tabular}}
\caption{Results on MEG and EEG dataset.}
\label{tab:meg_eeg}
\endgroup
\end{table}

\begin{figure}[ht]
\centering
\includegraphics[width=0.98\columnwidth]{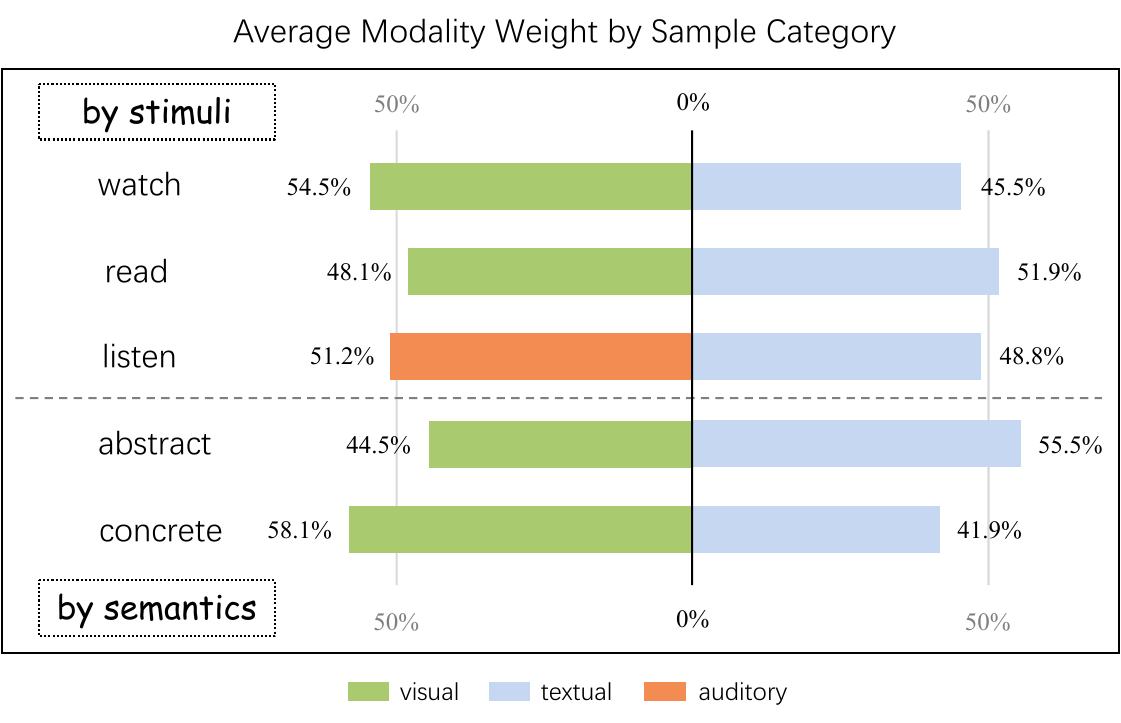}
\caption{Average Modality Router weights for visual, textual, and auditory embeddings across sample categories partitioned by stimulus type (watch/read/listen) and by semantic property (abstract/concrete).}
\label{fig:avg_modality}
\end{figure}

\subsection{Modality Router Analysis} \label{sec:analysis}

To understand why multimodal integration improves decoding, we partitioned samples by stimulus type and by semantic property, and analyzed modality-specific contributions via Modality Router weights. As shown in Figure \ref{fig:avg_modality}, router weights track the dominant input modality: watch-image samples emphasize visual embeddings, read-text samples emphasize textual embeddings, and listen-audio samples emphasize auditory embeddings. This mirrors neuroscientific evidence that modality-selective brain regions show stronger activation for matching sensory inputs \cite{regev2013selective,johnson2005attention}. Non-dominant modalities receive smaller but non-negligible weights, providing complementary signals. Note that the relatively small gaps between modalities arise because the load-balancing loss promotes cross-modality balance, preventing the router from collapsing into a single modality. 

Under the concreteness partition, textual weights increase for more abstract samples, whereas visual weights increase for more concrete samples, consistent with evidence that abstract concepts rely more on language systems and concrete concepts engage sensorimotor systems more strongly \cite{paivio1991dual}.

\begin{table}[t]
\centering
\begingroup
\setlength{\tabcolsep}{0.5mm}
\resizebox{\columnwidth}{!}{
\begin{tabular}{llcccc}
\toprule
Dataset & Modality & BLEU-1$\uparrow$ & ROUGE-1$\uparrow$ & ROUGE-L$\uparrow$ & WER$\downarrow$ \\
\midrule
\multirow{3}{*}{NSD}
  & visual & 59.57 & 44.24 & 44.04 & 73.39 \\
  & textual   & 59.24 & 43.81 & 43.65 & 74.10 \\
  & omni   & \textbf{61.53}  & \textbf{46.58} & \textbf{46.45} & \textbf{69.04} \\
\midrule
\multirow{3}{*}{Pereira}
  & visual     & 34.32 & 26.13 & 25.97 & 82.31 \\
  & textual       & 33.04 & 25.57 & 25.21 & 84.93 \\
  & omni       & \textbf{36.32}	&\textbf{28.31}	&\textbf{27.82}	&\textbf{79.32} \\
\midrule
\multirow{3}{*}{Huth}
  & auditory      & 15.31 & 13.19 & 13.13 & 95.42 \\
  & textual       & 16.01 & 14.02 & 14.03 & 93.87 \\
  & omni       & \textbf{17.35} & \textbf{15.89} & \textbf{15.28} & \textbf{91.35} \\
\bottomrule
\end{tabular}}
\caption{Ablation study of modalities utilized by the framework on NSD, Pereira, and Huth.}
\label{tab:modality_ablation}
\endgroup
\end{table}

\subsection{Modality Ablation Experiments} \label{sec:ablation}

To quantify the contribution of each modality and assess the necessity of multimodal fusion, we compared the full omni model with unimodal variants. As shown in Table \ref{tab:modality_ablation}, the omni approach consistently outperforms all unimodal variants across all datasets and evaluation metrics, achieving average relative improvements of 5.42\%, 7.81\%, and 10.96\% on NSD, Pereira, and Huth, respectively. These results demonstrate the effectiveness and robustness of multimodal fusion in capturing complementary information from different modalities.

Among unimodal counterparts, vision‐based models consistently lead in performance, likely because visual signals offer the richest, most intuitive alignment. In contrast, text‐based models trail behind, and audio‐based models lag the most, suggesting greater difficulty in mapping brain activity to auditory inputs. Additional analysis of the competency patterns of different modal counterparts is provided in Appendix \ref{app:modal_perf}.

\section{Conclusion}
This paper introduced a novel and generalizable BCI framework that embraces the multimodal nature of brain activity by aligning brain signals with multimodal representations from an MLLM across text, image, and audio domains. A key innovation is our router module, which adaptively selects and fuses modality-specific brain features, mirroring the brain's associative mechanisms. Extensive evaluations on three fMRI datasets achieved state-of-the-art performance, and the framework showed robustness when extended to EEG and MEG data, consistently achieving superior results across varying temporal and spatial resolutions. This approach brings us closer to real-world applications by enabling BCIs to process diverse input modalities, offering more accurate and flexible systems.

\section*{Limitations}
Although our results demonstrate the effectiveness of multimodal fusion and its consistency with the brain’s associative mechanisms, several limitations remain. First, the dimensionality reduction applied during data preprocessing limits our ability to establish precise correspondences between model channels and fine-grained brain regions, thereby constraining the interpretability of the learned representations. Preserving more spatial structure in future work may help improve neural interpretability and enable a more detailed analysis of brain–model alignment.

In addition, brain decoding technology is still largely at the laboratory stage and remains far from practical real-world deployment. In particular, achieving online real-time decoding will require coordinated advances across algorithms, software systems, and hardware platforms, rather than improvements in modeling alone. Future work could enhance the feasibility of online decoding through model compression and lightweight design, together with system-level optimization for efficient inference.

\section*{Ethical Statements}
This study used preprocessed data from publicly available datasets. The data were de-identified/anonymized as part of the original dataset releases and did not include direct personal identifiers. The original collection of the fMRI, MEG, and EEG data was conducted under appropriate ethical review and approval, as described in the corresponding dataset publications.

Brain decoding technologies, including the methods proposed in this study, have the potential for misuse if applied without appropriate safeguards. Possible concerns include violations of mental privacy, unauthorized inference of cognitive states, or deployment in coercive contexts. Although the present work focuses on methodological development within a controlled research setting and uses anonymized data under institutional ethical approval, we acknowledge the broader societal implications of brain decoding research. Future developments should be accompanied by clear regulatory frameworks, informed consent procedures, and robust data protection standards to ensure responsible use.

\bibliography{custom}

\appendix

\section{Dataset Details}
\label{app:dataset_details}

\begin{table*}[h]
\centering
{
\setlength{\aboverulesep}{4pt}
\setlength{\belowrulesep}{4pt}%
\begin{tabular}{lcccp{6cm}}
\toprule
Dataset   & Auxiliary Image & Auxiliary Audio & Stimuli & Text \\ \midrule
NSD
& \begin{minipage}[b]{0.3\columnwidth}
\centering
\raisebox{-.5\height}
{\includegraphics[width=2cm]{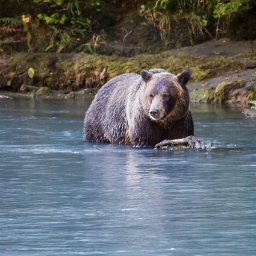}}
\end{minipage}
& \ding{55}       & Visual      & A bear that is standing in the water.\\ 

\midrule 
Pereira
& \begin{minipage}[b]{0.3\columnwidth}
\centering
\raisebox{-.5\height}
{\includegraphics[width=2cm]{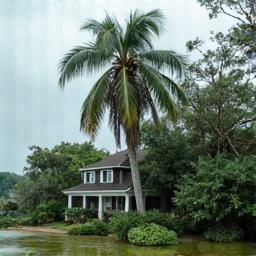}}
\end{minipage}
& \ding{51}       & Textual      & Hurricanes can tear limbs off of trees and suck houses off their foundations.\\ 

\midrule 
Huth
& \ding{55}
& \ding{51}       & Auditory      & Thank you very much this is my story...\\ 

\midrule 
SMN4Lang
& \ding{55}
& \ding{55}       & Auditory    & \begin{CJK}{UTF8}{gbsn}虽然已经过了处暑...\end{CJK}\\ 

\midrule 
ZuCo2
& \ding{55}
& \ding{55}       & Textual   & Henry Ford, with his son Edsel...\\ 

\bottomrule
\end{tabular}
}
\caption{Examples from each dataset.}
\label{tab:data_examples}
\end{table*}

For fMRI data, we employed three modality‐specific datasets evoked by visual (NSD; \citet{allen2022massive}), textual (Pereira; \citet{pereira2018toward}), and auditory stimuli (Huth; \citet{lebel2023natural}). For MEG data, we used the SMN4Lang dataset \cite{wang2022synchronized}, and for EEG data, the ZuCo 2 dataset \cite{zou2022cross}, details of all datasets are provided below. Examples from each dataset are provided in Table \ref{tab:data_examples}.

\subsection{Natural Scenes Dataset}
The Natural Scenes Dataset (NSD; \citet{allen2022massive}) is the largest publicly available human fMRI resource for visual neuroscience. It contains high-resolution brain activity recordings from eight participants, each of whom viewed a rich set of natural images inside a 7 T MRI scanner for up to 40 hours. Images from the MS COCO corpus were presented for three seconds and repeated three times in each session, resulting in approximately 22,000–30,000 trials per subject. Consistent with prior work, our analyses focus on the four participants (S1, S2, S5, and S7) who completed all scanning sessions. We also limit our analysis to voxels within the `nsdgeneral' region, which is defined by the NSD authors as the subset of posterior cortical sites most responsive to visual input. For these four subjects, the training set includes 8,859 unique images and 24,980 trials.

\subsection{Pereira Dataset}
The Pereira dataset \citep{pereira2018toward} comprises fMRI recordings acquired on a 3-Tesla Siemens Trio scanner with a 32-channel head coil, collected from five participants who completed two complementary reading experiments. Stimuli in both experiments were organized hierarchically-topic, passage, sentence-to probe semantic processing across varied contexts. In Experiment 1, participants read 96 Wikipedia-style passages, each composed of four sentences centered on one of 24 distinct concepts. Experiment 2 presented 72 passages—48 in the same encyclopedic format and 24 composed as first- or third-person narratives—each containing three to four sentences on concepts not repeated from Experiment 1. The experimental design ensured comparable levels of semantic relatedness both within and across passages and topics. Across both experiments, a total of 672 unique sentences served as stimuli; the resulting imaging data from all five subjects have been pooled to maximize statistical power in downstream analyses.

\subsection{Huth Dataset}

The Huth natural-language fMRI dataset \citep{lebel2023natural} is a naturalistic speech-perception dataset designed for voxelwise encoding and decoding studies. In the full public release, eight participants listened to 27 spoken narrative stories while BOLD fMRI signals were recorded, yielding a densely sampled single-subject dataset that is especially suitable for modeling language representations at the voxel level. Functional images were acquired with a repetition time of 2 s, and the dataset also includes a dedicated held-out test story, which was repeated across five scanning sessions to facilitate out-of-sample evaluation and estimates of response reliability. In addition to the fMRI data, the release provides time-aligned linguistic annotations, including word- and phoneme-level onset/offset boundaries, which make it possible to align stimulus-derived features precisely to the measured neural responses. Because of its relatively long recording duration per subject, fine-grained transcript alignment, and explicit train/test structure, this dataset has become a standard benchmark for studying how semantic and contextual information in continuous speech is represented across cortex.

\subsection{SMN4Lang Dataset}
The SMN4Lang MEG dataset comprises magnetoencephalography recordings from the same twelve participants who contributed to the fMRI collection \citep{wang2022synchronized}. Using a 306-sensor whole-head array, MEG captures neuronal activity with millisecond precision, albeit at coarser spatial resolution than fMRI. Participants listened to the identical set of sixty narrative stories employed in the fMRI experiment. To quantify word-level responses, the MEG signal following each word’s offset was segmented into nine successive 200 ms windows spanning the first second, with each window overlapping the previous by 100 ms. The mean amplitude within each window thus provides a fine-grained temporal profile of the brain’s response to every word in the discourse.

\subsection{ZuCo2}
The ZuCo2 dataset \citep{zou2022cross} comprises high‐temporal‐resolution electroencephalography (EEG) recordings from eighteen healthy adults (one of the original nineteen participants was excluded due to technical issues). During the experiment, each participant silently read a total of 739 sentences drawn from Wikipedia. The reading paradigm included two conditions: a Normal Reading task, in which subjects viewed 349 randomly selected Wikipedia sentences, and a Task‐Specific Reading task, in which they read 390 sentences organized around seven semantic categories-political affiliation, education, founder, spouse, job title, nationality, and employer. Continuous EEG signals were recorded throughout, capturing scalp voltage fluctuations at millisecond precision to support both general and category‐focused language processing analyses.

\section{Implementation details}

\subsection{Experimental Environment}
All experiments were implemented in PyTorch 2.6.0 and accelerated using CUDA 12.8. We leveraged the Hugging Face \texttt{transformers} library (v4.52.3) for model loading and tokenization. Model training was carried out on four NVIDIA Tesla V100 GPUs.

\subsection{Architecture and Training}
Each visual projector was initialized with a pre-trained CLIP ViT-L/14 backbone, while Qwen2.5-Omni 7B served as the auto-regressive language decoder. During the multimodal instruction tuning phase, we prepended 10 learnable soft prompt tokens to the textual input and employed 16 fixed-length learnable query vectors to interface with the visual features. Input images were resized to $112\times112\times3$ pixels and fed through the respective projectors. Progressive alignment was applied with a central time step of $t_0 = 1000$ and a transition sharpness controlled by $\lambda = \ln(999)$.

In the subsequent projector fusion phase, we balanced the objectives by setting both the alignment loss weight and the language modeling loss weight to 1. A load‐balancing regularization term was included with a weight of 0.01.

Optimization was performed using the AdamW optimizer with parameters $\beta_1=0.9$, $\beta_2=0.999$, and a fixed learning rate of $5\times10^{-5}$. A batch size of 32 was maintained throughout both training phases. For auxiliary image generation during training, we employed FLUX.1-dev with a guidance scale of 3.5 to augment the visual signal. We use greedy decoding for text generation.

\subsection{Statistical Significance}

We evaluated the statistical significance of our results using the Student's t-test. The proposed method was compared against all established baselines, with all experiments conducted across 10 independent runs using distinct random seeds. All reported results are expressed as mean values. Our experimental results demonstrate that the proposed method achieves statistically significant improvements across all benchmarks (with all p-values < 0.05).

\begin{table*}[ht]
  \centering
  \label{tab:method-comparison}
  \resizebox{\textwidth}{!}{
  \begin{tabular}{l|ccc}
    \toprule
    Method & Multimodal Stimuli & Multimodal Brain & Multimodal Mind \\
    \midrule
    SDRecon \citep{takagi2023high}     & \ding{55}  & \ding{55}   & \ding{55}  \\
    OneLLM \citep{han2024onellm}     & \ding{55}  & \ding{55}   & \ding{55}  \\
    UniBrain \citep{mai2023unibrain}    & \ding{55}  & \ding{55}   & \ding{55}  \\
    BrainCap \citep{ferrante2023brain}    & \ding{55}  & \ding{55}   & \ding{55}  \\
    BrainChat \citep{huang2025brainchat}   & \ding{55}  & \ding{55}   & \ding{55}  \\
    UMBRAE \citep{xia2024umbrae}      & \ding{55}  & \ding{55}   & \ding{55}  \\
    MindLLM \citep{qiu2025mindllm}     & \ding{55}  & \ding{55}   & \ding{55}  \\
    BrainLLM \citep{ye2025generative}   & \ding{51}  & \ding{55}   & \ding{55}  \\
    Ours        & \ding{51}  & \ding{51}   & \ding{51}  \\
    \bottomrule
  \end{tabular}}
  \caption{Comparison of our model against existing methods. \textit{Multimodal stimuli} refers to the variety of external inputs (e.g., images, sounds, text) presented during brain‐signal recording. \textit{Multimodal brain} denotes the different neuroimaging modalities employed (e.g., fMRI, MEG, EEG). \textit{Multimodal mind} indicates the internal perceptual or conceptual reconstructions evoked by those stimuli (for example, the mental image and sound of a barking dog triggered by text describing a dog).}
  \label{tab:method_comparison}
\end{table*}

\begin{table*}[ht]
\centering
\begingroup
\setlength{\tabcolsep}{1.1mm} 
\resizebox{\textwidth}{!}{
\begin{tabular}{@{}lcccccccccc@{}}
\toprule
Strategy & Subject & BLEU-1 & BLEU-2 & BLEU-3 & BLEU-4 & METEOR & ROUGE-L & CIDEr & CLIP-S & RefCLIP-S \\
\midrule
\multirow{4}{*}{soft merge} & 1 & 61.46 & 43.37 & 30.07 & 21.04 & 20.60 & 46.15 & 63.68 & 62.72 & 69.72 \\
& 2 & 61.38 & 43.30 & 30.00 & 20.99 & 20.53 & 46.07 & 63.61 & 62.65 & 69.66 \\
& 5 & 62.45 & 44.36 & 31.06 & 22.03 & 21.59 & 47.14 & 64.67 & 63.71 & 70.71 \\
& 7 & 59.47 & 41.38 & 28.08 & 19.05 & 18.61 & 44.16 & 61.69 & 60.73 & 67.73 \\
\midrule
\multirow{4}{*}{hard select} & 1 & 60.69 & 42.59 & 29.90 & 21.33 & 19.67 & 45.66 & 59.55 & 60.31 & 67.75 \\
& 2 & 60.61 & 42.52 & 29.83 & 21.28 & 19.60 & 45.58 & 59.48 & 60.24 & 67.69 \\
& 5 & 61.68 & 43.58 & 30.89 & 22.32 & 20.66 & 46.65 & 60.54 & 61.30 & 68.74 \\
& 7 & 58.70 & 40.60 & 27.91 & 19.34 & 17.68 & 43.67 & 57.56 & 58.32 & 65.76 \\
\midrule
\multirow{4}{*}{similarity merge} & 1 & 61.53 & 43.45 & 30.21 & 21.92 & 20.71 & 46.45 & 63.88 & 63.13 & 70.12 \\
& 2 & 61.47 & 43.38 & 30.15 & 21.89 & 20.65 & 46.39 & 63.81 & 63.07 & 70.05 \\
& 5 & 62.59 & 44.51 & 31.28 & 22.98 & 21.75 & 47.50 & 64.92 & 64.18 & 71.19 \\
& 7 & 59.50 & 41.40 & 28.18 & 19.85 & 18.68 & 44.42 & 61.85 & 61.10 & 68.08 \\
\bottomrule
\end{tabular}}
\caption{Results of brain captioning on NSD subject 1, 2, 5, 7 using three different routing strategy.}
\label{tab:other_subjects}
\endgroup
\end{table*}

\section{Method Comparison}
\label{app:method_comp}
As summarized in Table \ref{tab:method_comparison}, our approach is the first to jointly address all three key dimensions of multimodal brain decoding. While earlier methods such as SDRecon \citep{takagi2023high}, OneLLM \citep{han2024onellm}, UniBrain \citep{mai2023unibrain}, BrainCap \citep{ferrante2023brain}, BrainChat \citep{huang2025brainchat}, UMBRAE \citep{xia2024umbrae}, and MindLLM \citep{qiu2025mindllm} focus exclusively on a single data modality (neural signals) and do not integrate varied external stimuli or reconstruct internal representations, BrainLLM \citep{ye2025generative} extends support to multimodal stimuli but still lacks the ability to model diverse brain imaging modalities and mental representations. In contrast, our model simultaneously ingests heterogeneous stimuli (images, sounds, text), fuses multiple neuroimaging modalities (e.g., fMRI, MEG, EEG), and reconstructs the corresponding perceptual and conceptual contents in the mind. This unified treatment enables richer, more flexible decoding of human cognition than any prior work.

\section{Experiments on Other Subjects}
\label{app:other_sub}
To assess the robustness of our three routing strategies beyond the primary subject (NSD subject 1), we conducted additional brain‐captioning experiments on three independent NSD subjects (subject 2, 5, and 7). Table \ref{tab:other_subjects} summarizes a variety of standard captioning metrics—BLEU‐1 through BLEU‐4 \citep{papineni2002bleu}, METEOR \citep{banerjee2005meteor}, ROUGE‐L \citep{lin2004rouge}, CIDEr \citep{vedantam2015cider}, CLIP‐S and RefCLIP‐S \citep{hessel2021clipscore}—for each strategy.

Across all subjects, \emph{similarity merge} consistently achieves the highest or near‐highest scores on nearly every metric, indicating superior caption fluency and semantic alignment with the reference images. In particular, on subject 5 it yields the best absolute results (e.g., BLEU‐1: 62.59, CIDEr: 64.92, RefCLIP‐S: 71.19), demonstrating its robust generalization to unseen brain patterns. \emph{Soft merge} closely follows, outperforming \emph{hard select} by a clear margin in terms of n-gram overlap (BLEU) and embedding‐based metrics (CLIP‐S). \emph{Hard select}, while simpler, lags behind on both lexical accuracy and conceptual relevance, suggesting that the all-or-nothing routing decision may discard useful multimodal cues.

These results confirm that our routing strategies maintain their relative ordering of effectiveness across multiple subjects, with \emph{similarity merge} emerging as the most reliable mechanism for integrating visual and neural representations in brain‐captioning tasks.

\section{Additional Analysis}

\begin{figure*}[h]
\centering
\includegraphics[width=0.8\textwidth]{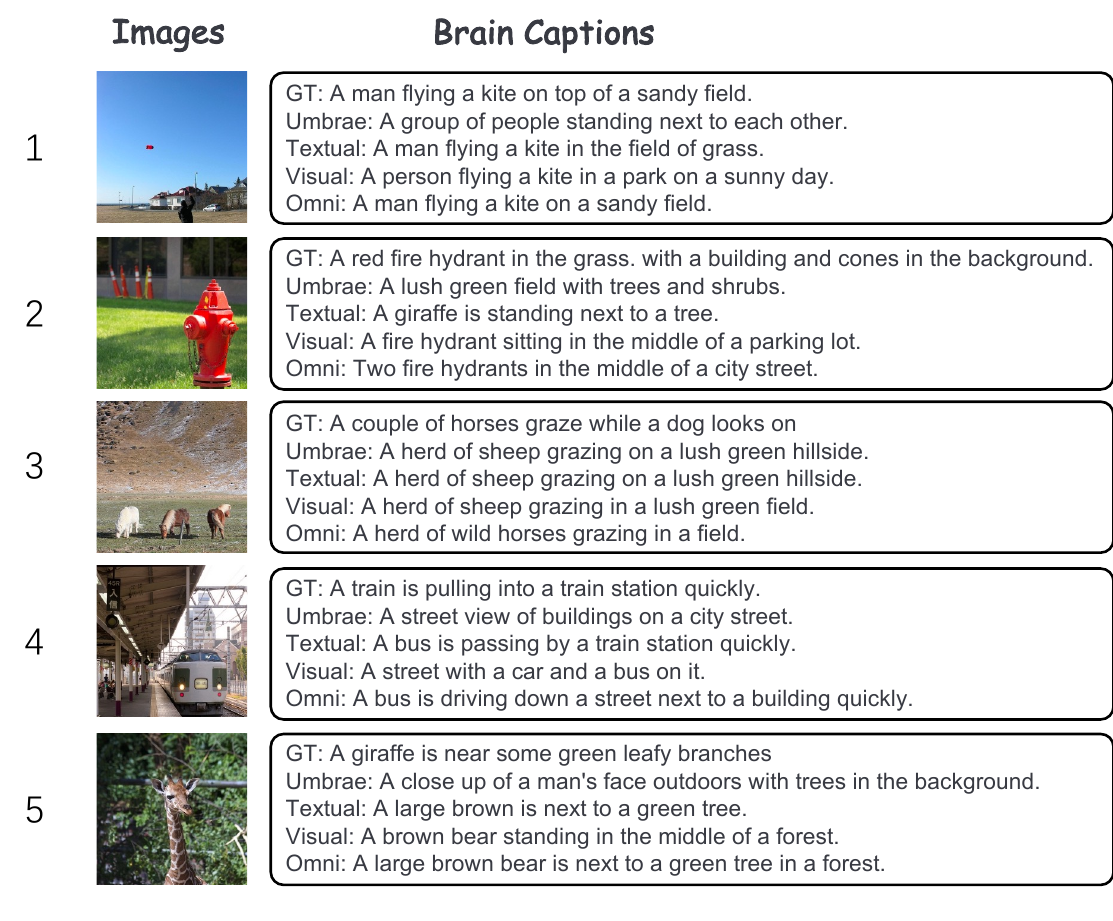} 
\caption{Decoding results on the NSD for the models: Umbrae, textual, visual, Omni, and the ground truth.}
\label{decode_res}
\end{figure*}

\begin{table}[t]
\centering
\begingroup
\setlength{\tabcolsep}{0.9mm}
\resizebox{\columnwidth}{!}{
\begin{tabular}{lccc}
\toprule
Method & BLEU-1 & ROUGE-L & CLIP-S \\
\midrule
Full model & 61.53 & 46.45 & 63.13 \\
\phantom{..}w/o load balancing loss & 52.29 & 38.18 & 60.15 \\
\phantom{..}w/ ~ KL alignment & 59.03 & 45.19 & 60.83 \\
\phantom{..}w/o progressive alignment & 58.78 & 44.74 & 60.01 \\
\phantom{..}w/o phase 1 & 40.21 & 27.39 & 57.04 \\
\phantom{..}w/o phase 2 & 36.13 & 24.30 & 60.38 \\
\phantom{..}w/o soft prompts & 60.02 & 45.93 & 61.59 \\
\phantom{..}w/o soft prompt text init & 60.23 & 46.19 & 62.04 \\
\bottomrule
\end{tabular}}
\caption{Ablation study on model components and training strategies on NSD.}
\label{tab:other_ablation}
\endgroup
\end{table}

\subsection{Ablation on Model Components and Training Strategies}
We evaluate the contributions of four essential components—load balancing loss, alignment strategy, two-phase training, and prompt tuning—through ablation studies, as presented in Table \ref{tab:other_ablation}. The removal of load balancing loss results in a substantial performance decline, demonstrating its critical role in balancing projector activation. With respect to alignment strategies, we observe that progressive alignment using MSE outperforms KL-based alignment, and omitting alignment altogether leads to further performance degradation. Furthermore, the two-phase training framework is indispensable; eliminating either phase leads to dramatic reductions across all performance metrics, emphasizing the necessity of both stages. Lastly, prompt tuning, especially with soft prompts and text-based initialization, yields modest improvements. Although beneficial, prompt tuning contributes less to overall performance compared to the other components.

\begin{table}[t]
    \centering
    \begin{tabular}{lcc}
        \toprule
        method & Grammar & CLIP-S \\
        \midrule
        textual & 74.78 & 59.63 \\
        visual & 72.46 & 61.92 \\
        omni & 74.81 & 63.13 \\
        \bottomrule
    \end{tabular}
    \caption{The text and image performance of different multimodal models. \textit{Grammar} measures the textual grammatical capability of the model, while \textit{CLIP-S} evaluates the correlation between the generated text and images.}
\label{tab:text_image}
\end{table}

\subsection{Discrepancy in Modality Performance}
\label{app:modal_perf}
To examine the relative strengths of different modal counterparts and to better understand why multimodal fusion is effective, we evaluate their textual grammaticality and text–image alignment, as reported in Table \ref{tab:text_image}. The textual model aligns only with the textual brain representation and therefore achieves stronger performance on the Grammar metric. In contrast, the visual model aligns solely with the visual representation, resulting in higher text–image correlation as measured by CLIP-S. By simultaneously aligning with both textual and visual representations, the Omni model attains the best results on both Grammar and CLIP-S, indicating that integrating the two modalities improves both grammatical quality and cross-modal consistency.

\subsection{Decoding results}
\label{app:examples}
Figure \ref{decode_res} presents the ground truth for brain captioning along with the results from four models: Umbrae \citep{xia2024umbrae}, textual, visual, and Omni. The textual model closely mirrors the original captions in terms of grammar and sentence structure (1, 2). The visual model excels at describing the content of the images accurately (1, 4). The Omni model, which combines the linguistic reconstruction of the textual model with the image encoding capability of the visual model, delivers the best performance overall (1, 3, 4). While some cases show slight differences between the decoded results and the original captions, the subjects and the original sentences are more closely aligned in comparison to the Umbrae model (5).

\section{Licenses of scientific artifacts}

We follow and report the licenses of scientific artifacts involved in Table \ref{tab:license}.

\begin{table}[ht]
 \setlength{\tabcolsep}{6mm}
 \centering
 \small
 \renewcommand\arraystretch{1.25}
 \begin{center}
  \begin{tabular}{ll}
   \toprule[1.2pt]  
               \multicolumn{1}{l}{\textbf{Name}} & \multicolumn{1}{l}{\textbf{License}} \\
                \midrule[0.8pt]
                Transformers      & Apache 2.0 license     \\
                Qwen2.5-Omni-7B   & Apache 2.0 license \\
                clip-vit & MIT license \\
                NSD & No License \\
                Pereira & No license \\
                Huth & CC0 \\
                SMN4Lang & CC0 \\
                ZuCo2 & No License \\
   \bottomrule[1.2pt]
  \end{tabular}
 \end{center}
    \caption{\label{tab:license} Licenses of scientific artifacts involved in this work.}
\end{table}


\end{document}